\documentclass[11pt]{article}
\usepackage[utf8]{inputenc}
\usepackage[T1]{fontenc}
\usepackage{lmodern}
\usepackage{amsmath,amssymb,amsthm,bm}
\usepackage{booktabs,longtable,array,multirow,graphicx,geometry,hyperref,microtype,calc}
\usepackage{caption}
\usepackage{authblk}
\usepackage{enumitem}
\usepackage{float}
\graphicspath{{media/media/}}
\hypersetup{colorlinks=true,citecolor=blue,linkcolor=blue,urlcolor=blue}
\newtheorem{definition}{Definition}
\newtheorem{lemma}{Lemma}
\newtheorem{theorem}{Theorem}
\newcommand{\CVaR}{\operatorname{CVaR}}
\title{Robust Market Making with Hawkes Order Flow and Price Impact via Adversarial Reinforcement Learning}

\author[1]{Hao Yang\thanks{Corresponding author}}
\author[2]{Zhenguo Xu}

\affil[1]{
North China Institute of Computer System Engineering, Beijing, China\\
\texttt{yanghao244@mails.ucas.edu.cn}
}

\affil[2]{
University of Science and Technology of China, Hefei, China\\
\texttt{xzg@mail.ustc.edu.cn}
}

\date{}

\begin{document}
\maketitle

\begin{abstract}
Market-making strategies in real limit-order-book markets face substantial model uncertainty and regime-shift risk. Prior work improves robustness to model misspecification by introducing an environmental adversary and reformulating the Avellaneda--Stoikov market-making model as a zero-sum game between the market maker and the market environment. Existing approaches, however, are largely built on simplified assumptions of Poisson order arrivals and no price impact, which makes it difficult to represent high-frequency microstructure effects such as clustered order flow, self-excitation, and post-trade price feedback. We extend adversarial reinforcement learning (ARL) to a richer market-making environment with Hawkes self-exciting arrivals and trade-induced price impact, and introduce an LSTM module to mitigate the stronger non-stationarity created by this extension. We further characterize the equilibrium properties of the proposed framework from both game-theoretic and numerical perspectives, and propose a robustness evaluation protocol focused on left-tail improvement to match the worst-case nature of ARL training. Experiments show that the proposed method achieves better left-tail return performance in most complex microstructure environments, with no evidence that the improvement is driven by stronger terminal directional inventory bias.
\end{abstract}


\section{Introduction}
Market making refers to the provision of liquidity by continuously posting bid and ask quotes for a financial asset. Market makers earn the bid--ask spread, but are simultaneously exposed to inventory risk, adverse-selection risk, and model-misspecification risk. Traditional market-making models often assume that the market maker knows the market environment, including the price process, order-arrival intensity, and execution probabilities. In real limit-order-book markets, however, liquidity, order-flow clustering, price impact, and drift regimes vary over time and are often only partially observed. The Avellaneda--Stoikov model is a classical control framework for market making~\cite{avellaneda2008}. It assumes Brownian mid-price dynamics and Poisson market-order arrivals on the bid and ask sides, yielding an analytically tractable framework for optimal quotes under inventory risk. Subsequent optimal-control studies have extended this framework to alternative utility functions, risk preferences, and market settings~\cite{fodra2012,gueant2013,cartea2015}. Recent work has studied robustness from several directions, including robust control under model uncertainty~\cite{cartea2017}, adversarial training~\cite{spooner2020,pinto2017,rajeswaran2017}, richer action spaces~\cite{wang2023}, and non-Markov market-state modeling~\cite{lalor2025}.

This paper focuses on adversarial reinforcement learning. The market-making problem is cast as a zero-sum game, and the agents are trained toward an approximate Nash equilibrium so that the market maker learns a conservative policy that limits losses across adverse environments. Robust Market Making via Adversarial Reinforcement Learning (ARLMM) converts the Avellaneda--Stoikov model into a zero-sum stochastic game between a market maker and an environmental adversary~\cite{spooner2020}. The adversary chooses market parameters to minimize the market maker's cumulative return, while the market maker learns a quoting strategy that remains effective under these unfavorable regimes. Compared with fixed-environment and randomly sampled-environment training, ARLMM can improve robustness to model misspecification. Its market microstructure, however, remains relatively simple: order arrivals are Poisson and executions do not directly feed back into the mid-price. These assumptions omit two important high-frequency effects. First, real order flow is often clustered and self-exciting, meaning that one trade can increase the short-run probability of subsequent trades; Hawkes processes provide a natural representation of this effect~\cite{hawkes1971,bacry2015}. Second, executions may themselves affect short-run prices, feeding trading activity back into inventory valuation and mark-to-market returns~\cite{almgren2001}.

We therefore extend ARLMM to an adversarial market-making environment with Hawkes order arrivals and trade-induced price impact. In ARLMM, the environmental adversary chooses a regime parameter vector $(b,A,k)$ at the start of each episode. We expand this vector to $(b,A,k,\kappa_H,\gamma_H,\xi)$, where $\kappa_H$ and $\gamma_H$ control Hawkes dynamics and $\xi$ controls trade-induced price impact. Table~\ref{tab:microstructure} summarizes the extension.

\begin{table}[H]
\centering
\caption{Comparison of market-microstructure mechanisms.}
\label{tab:microstructure}
\begin{tabular}{lcc}
\toprule
Market dynamic / mechanism & ARLMM & Proposed method \\
\midrule
Mid-price drift ($b$) & Yes & Yes \\
Baseline liquidity ($A$) & Yes & Yes \\
Order-book depth / execution sensitivity ($k$) & Yes & Yes \\
Poisson arrivals & Yes & Special case \\
Hawkes self-exciting arrivals $(\kappa_H,\gamma_H)$ & No & Yes \\
Trade-induced price impact $(\xi)$ & No & Yes \\
Trade-to-arrival-intensity feedback & No & Yes \\
Immediate trade-to-mid-price feedback & No & Yes \\
\bottomrule
\end{tabular}
\end{table}

The main challenge created by this extension is stronger non-stationarity. ARLMM already contains an adaptive regime-selection mechanism; enlarging the adversary's parameter space makes it easier for training to concentrate on a small number of adverse regimes, which can amplify non-stationarity from the market maker's perspective. In our experiments, this effect is especially visible in three classes of environments: Hawkes-dominated environments, environments combining high drift with high execution pressure, and environments dominated by liquidity structure. Because these environments contain risk patterns that propagate through time, we introduce an LSTM over the observation window to improve regime identification. We refer to this method as \emph{Ours}.

We also adopt a robustness evaluation protocol more closely aligned with worst-case training. Because ARLMM deliberately selects unfavorable markets at the episode level, a left-tail statistic is more informative than a metric such as the Sharpe ratio that mixes both tails of the return distribution. We therefore use $\CVaR_{10\%}$ and $\CVaR_{30\%}$, with $\CVaR_{10\%}$ as the primary metric~\cite{rockafellar2000}. To rule out the possibility that higher return is obtained through a larger directional inventory exposure, we additionally test terminal inventory bias across the 19 evaluation environments. Under our operational definition, a method is more robust when it has higher $\CVaR_{10\%}$ and there is no evidence that the improvement is purchased by stronger terminal directional inventory bias.

\subsection{Main Contributions}
The paper makes three contributions. First, we extend adversarial market-making training to a richer microstructure environment with Hawkes arrivals and price impact, and introduce an LSTM state representation to address the stronger temporal non-stationarity. Second, we characterize equilibrium properties at two levels: we establish equilibrium existence for a reduced single-stage game and examine approximate equilibrium behavior empirically in the multi-stage setting. Third, we propose a stricter robustness protocol that combines left-tail return metrics with a one-sided bootstrap test of terminal inventory bias.

\subsection{Related Work}
\paragraph{Optimal control and microstructure-based market making.} The theoretical study of market making originates in classical work on bid--ask spreads, inventory control, and information asymmetry~\cite{ho1981,glosten1985,grossman1988}. Avellaneda and Stoikov~\cite{avellaneda2008} combined mid-price diffusion, Poisson order arrivals, and inventory risk in a tractable limit-order-book model. Subsequent research has extended utility functions, inventory penalties, multi-asset settings, and general stochastic-control formulations~\cite{fodra2012,gueant2013,cartea2015,gueant2017}; studied robust market making under model misspecification~\cite{cartea2017}; and incorporated richer microstructure mechanisms such as Hawkes processes for persistent order flow~\cite{hawkes1971,bacry2015,abergel2013,morariu2022,jusselin2021} and models of price impact~\cite{almgren2001,fujii2015,singh2021}. Our work retains the modeled-market and zero-sum-game structure of ARLMM while extending arrivals from Poisson to Hawkes dynamics and adding trade-induced price impact.

\paragraph{Machine learning for market making.} Reinforcement learning has been used to learn quoting policies directly, including early work on electronic market makers~\cite{chan2001} and inventory-sensitive agents in data-driven limit-order-book environments~\cite{spooner2018}. Online and offline learning methods have been explored to improve adaptivity, stability, and sample efficiency~\cite{abernethy2013}. Reinforcement learning has also been applied to optimal execution and risk-sensitive decision making~\cite{nevmyvaka2006,vyetrenko2019}. Deep learning has been used to extract high-dimensional market states and improve policy representations, including model-based methods for multi-asset market making~\cite{gueant2019}. Model-based environments such as mbt-gym provide controlled and reproducible testbeds for comparing market-making algorithms~\cite{jerome2022}. Our emphasis is on temporal state representation within adversarially trained market-making policies.

\paragraph{Robustness and risk-sensitive reinforcement learning.} Robustness to model uncertainty has long been studied in algorithmic trading~\cite{cartea2017}. In reinforcement learning, related work includes policy search under risk criteria~\cite{tamar2012}, distributional reinforcement learning~\cite{bellemare2017}, safe reinforcement learning~\cite{garcia2015}, and particle-based value estimation~\cite{maddison2017}. A complementary line of work uses adversarial training to handle environmental perturbations and model uncertainty, including robust adversarial reinforcement learning~\cite{pinto2017}, actor--critic fictitious play in multi-stage games~\cite{perolat2018}, and robust policy learning with model ensembles~\cite{rajeswaran2017}. ARLMM~\cite{spooner2020} brings this adversarial-training idea to market making and provides the direct methodological foundation for our work.

\section{Trading Model}
We consider single-asset market making. Let $Z_n$ denote the mid-price, $H_n$ inventory, and $X_n$ cash. At each discrete time step, the market maker posts one-unit bid and ask orders around the mid-price. With trade-induced price impact, the mid-price follows
\begin{equation}
Z_{n+1}=Z_n+b_n\Delta t+\sigma_n W_n+\xi_n(\Delta N_n^- - \Delta N_n^+),
\label{eq:midprice}
\end{equation}
where $b_n$ is drift, $\sigma_n$ is volatility, $W_n$ is an independent Gaussian shock with mean zero and variance $\Delta t$, and $\xi_n\ge 0$ is the price impact induced by one unit of net execution. Let $p_n^+$ and $p_n^-$ denote the bid and ask quotes, respectively. Quote depths are
\begin{equation}
\delta_n^{\pm}=\pm(Z_n-p_n^{\pm})\ge 0.
\label{eq:depth}
\end{equation}
The spread and reservation price are
\begin{equation}
\psi_n=\delta_n^++\delta_n^-,\qquad p_n=\frac12(p_n^++p_n^-)=Z_n+\frac12(\delta_n^- - \delta_n^+).
\label{eq:spread}
\end{equation}

We use bilateral Hawkes conditional intensities $\lambda_n^{\pm}$ with the discrete-time dynamics
\begin{equation}
\lambda_{n+1}^{\pm}=\lambda_n^{\pm}+\kappa_{H,n}(A_n-\lambda_n^{\pm})\Delta t+\gamma_{H,n}\Delta N_n^{\pm},
\label{eq:hawkes}
\end{equation}
where $A_n>0$ is baseline liquidity, $\kappa_{H,n}>0$ controls mean reversion toward the baseline, and $\gamma_{H,n}\ge 0$ controls self-excitation. Conditional on quote depth, the effective execution intensity is
\begin{equation}
\Lambda_n^{\pm}=\lambda_n^{\pm}e^{-k_n^{\pm}\delta_n^{\pm}},
\label{eq:effective_intensity}
\end{equation}
where $k_n^{\pm}>0$ describes depth decay. The Poisson model is recovered as the special case $\gamma_{H,n}=0$ and $\lambda_n^{\pm}\equiv A_n$.

Inventory evolves according to
\begin{equation}
H_{n+1}=H_n+\Delta N_n^+-\Delta N_n^-.
\label{eq:inventory}
\end{equation}
The cash process can be written as
\begin{equation}
X_{n+1}=X_n+p_n^-\Delta N_n^- - p_n^+\Delta N_n^+
= X_n+\delta_n^-\Delta N_n^-+\delta_n^+\Delta N_n^+-Z_n\Delta H_n.
\label{eq:cash}
\end{equation}
The mark-to-market portfolio value is
\begin{equation}
\Pi(X,H,Z)=X+HZ.
\label{eq:wealth}
\end{equation}
Equations~\eqref{eq:midprice}--\eqref{eq:wealth} define the market-making dynamics used below. The single-stage game in Section~\ref{sec:game} conditions on the current state $(H_n,\lambda_n^+,\lambda_n^-)$, so the Hawkes extension enters the one-stage payoff through current conditional intensities. In the multi-stage experiments, the environmental adversary selects a regime $(b,A,k,\kappa_H,\gamma_H,\xi)$ that induces the corresponding price, arrival, and execution paths throughout an episode. The reinforcement-learning reward uses increments in $\Pi$ together with inventory penalties.

\section{Game Formulation and Single-Stage Analysis}
\label{sec:game}
We define a zero-sum stochastic game between a market maker and an adversary, where the adversary may be interpreted as a proxy for the rest of the market.

\begin{definition}[Extended Market-Making Game]
The extended market-making game has $N$ stages. At each stage, the market maker chooses quote depths $\delta^{\pm}$ and the environmental adversary chooses $(b,A,k,\kappa_H,\gamma_H,\xi)$. The market maker's stage payoff is the expected change in mark-to-market value, $\mathbb{E}[\Delta\Pi]$, and the adversary receives its negative. The total payoff is the sum of stage payoffs.
\end{definition}

We first study the single-stage case $N=1$; Sections~\ref{sec:training} and~\ref{sec:experiments} then analyze the $N=200$ setting empirically. Once current inventory $H$ and the current Hawkes state are fixed, the effects of $(A,\kappa_H,\gamma_H)$ on the current-period payoff are summarized by $(\lambda^+,\lambda^-)$. Conditioning on $(H,\lambda^+,\lambda^-)$ and omitting the common factor $\Delta t$, Equations~\eqref{eq:midprice}--\eqref{eq:wealth} yield
\begin{equation}
f(\delta^{\pm};b,\lambda^{\pm},k^{\pm},\xi)
=\Lambda^+(\delta^+ + b-\xi)+\Lambda^-(\delta^- - b-\xi)+bH,
\label{eq:singlepayoff}
\end{equation}
where $\Lambda^{\pm}=\lambda^{\pm}e^{-k^{\pm}\delta^{\pm}}$.

\begin{lemma}[Payoff concavity in $\delta^{\pm}$]
For fixed $(b,\lambda^{\pm},k^{\pm},\xi)$, the payoff $f$ is concave in $\delta^+$ and $\delta^-$ on
\begin{equation}
\left[0,\frac{2}{k^+}-b+\xi\right],\qquad
\left[0,\frac{2}{k^-}+b+\xi\right],
\label{eq:concavity_interval}
\end{equation}
respectively.
\end{lemma}

\begin{theorem}[Nash equilibrium for fixed $(\lambda^{\pm},k^{\pm},\xi)$]
Let $b\in[\underline b,\overline b]$ with finite endpoints. Then the single-stage game admits a pure-strategy Nash equilibrium $(\delta^+,\delta^-,\widehat b)$. If the equilibrium is interior, the market maker's best response satisfies
\begin{equation}
\delta^+=\frac{1}{k^+}-b+\xi,\qquad
\delta^-=\frac{1}{k^-}+b+\xi,
\label{eq:bestresponse}
\end{equation}
and
\begin{equation}
\widehat b\in\arg\min_{b\in[\underline b,\overline b]}
 f(\delta^+(b),\delta^-(b);b,\lambda^{\pm},k^{\pm},\xi).
\end{equation}
\end{theorem}
A proof is given in Appendix~\ref{app:theorem1}.

\begin{theorem}[Nash equilibrium for the general reduced-form game]
Suppose $(b,\lambda^+,\lambda^-,\xi)$ ranges over a Cartesian product of bounded closed intervals and $k^{\pm}>0$ is fixed. Then the reduced-form single-stage game admits a mixed-strategy Nash equilibrium.
\end{theorem}
A proof is given in Appendix~\ref{app:theorem2}.

\section{Adversarial Training}
\label{sec:training}
The single-stage analysis is informative but does not capture the temporal structure of the full problem. We therefore consider a multi-stage setting and compare four training environments with increasing adversarial control over market dynamics.

\paragraph{Fixed.} The environment always uses
\[
(b,A,k,\kappa_H,\gamma_H,\xi)=(0,140,1.5,60,0,0),
\]
which corresponds to the standard drift, liquidity, and depth setting with Hawkes self-excitation and price impact disabled. This is a single-agent reinforcement-learning problem with stationary transitions.

\paragraph{Random.} At the start of each episode, parameters are sampled independently and then held fixed throughout the episode:
\[
b\in[-5,5],\quad A\in[105,175],\quad k\in[1.125,1.875],\quad
\kappa_H\in[35,60],\quad \gamma_H\in[0,40],\quad \xi\in[0,1].
\]
Each dimension is sampled from a truncated normal distribution centered at the standard environment. This setting captures passive robustness to model mismatch under non-stationary episode-level transitions.

\paragraph{ARLMM.} The third environment introduces a learnable zero-sum adversary following Robust Market Making via Adversarial Reinforcement Learning~\cite{spooner2020}. At the episode level, the adversary selects $(b_n,A_n,k_n)$ within the same ranges used by Random.

\paragraph{Ours.} We extend the ARLMM zero-sum training framework to the Hawkes-and-impact environment. An LSTM encodes the time-ordered observation window to improve sensitivity to liquidity clustering and impact regimes. The adversary selects $(b_n,A_n,k_n,\kappa_H,\gamma_H,\xi)$ at the episode level, with ranges identical to Random.

\subsection{Learning Configuration}
For ARLMM and Ours, both agents use Proximal Policy Optimization (PPO), an actor--critic method that stabilizes optimization by limiting the size of policy updates~\cite{schulman2017}. The policy network outputs distribution parameters for bid- and ask-side quote actions, while the value network estimates expected returns and generalized advantage estimation is used for policy gradients.

\paragraph{States.} The market maker's observation at time $t$ is $o_t^{MM}=(H_t,t)$, where $H_t$ is current inventory. All methods use a history window of the most recent $K=40$ observations,
\[
O_t^{MM}=(o_{t-K+1}^{MM},\ldots,o_t^{MM}).
\]
The environment's full internal state contains price, inventory, cash, and bilateral conditional intensities, but only inventory and time are exposed to the policy. For Ours, the observation window is encoded by a one-layer LSTM with 32 hidden units, $\widehat O_t^{MM}=\operatorname{LSTM}(O_t^{MM})$.

\paragraph{Policies.} The market maker outputs the continuous action $a_t^{MM}=(\delta_t^b,\delta_t^a)$, corresponding to bid- and ask-side quote offsets. The environmental adversary policy $\pi_{\psi}^{Adv}$ selects either $(b_n,A_n,k_n)$ for ARLMM or $(b_n,A_n,k_n,\kappa_H,\gamma_H,\xi)$ for Ours at the episode level.

\paragraph{Rewards.} The market maker's one-step reward is
\begin{equation}
r_t^{MM}=\Delta\Pi_t-\zeta H_t^2-\mathbf{1}_{\{t=T\}}\eta H_T^2,
\label{eq:reward}
\end{equation}
with $\zeta=10^{-4}$ and $\eta=10^{-2}$. Episode return is $R^{MM}=\sum_t r_t^{MM}$, and the adversary receives $R^{Adv}=-R^{MM}$.

\section{Experiments}
\label{sec:experiments}
In all experiments, actor and critic networks are multilayer perceptrons with one hidden layer of 64 units. The learning rate is $3\times10^{-4}$ and the discount factor is $0.999$. The horizon is $T=1$ with $\Delta t=0.005$, giving 200 discrete steps per trajectory. Each episode contains 50 trajectories; an episode is used as one training chunk, with 10 optimization epochs and a total budget of 800,000 environment steps. The mid-price diffusion coefficient is $\sigma=2$, the initial price is 100, and the inventory bound is 50.

Evaluation uses 19 test environments covering the standard environment, upward and downward drift, sparse and dense liquidity, Hawkes-only regimes, impact-only regimes, and several extreme compound-stress regimes. Results are averaged over five random seeds: 701, 30, 90, 45, and 78.

\subsection{Robustness Metrics}
ARLMM is a worst-case training procedure: at the episode level, the environmental player actively selects unfavorable markets to expose weaknesses in the market-making policy. For this reason, left-tail metrics are more aligned with the training objective than the Sharpe ratio. We use $\CVaR_{10\%}$ and $\CVaR_{30\%}$, taking $\CVaR_{10\%}$ as the primary metric~\cite{rockafellar2000}.

To test whether higher returns are obtained through stronger directional inventory exposure, we construct a nonparametric one-sided bootstrap test for terminal-inventory mean shift. For method $m$, let terminal-inventory samples be $\{Q_T^{(i)}\}_{i=1}^n$ and define
\[
T_m=\left|\widehat{\mathbb E}_m[Q_T]\right|.
\]
For Ours and a reference method Ref (Random or ARLMM), use
\begin{equation}
D_{\mathrm{obs}}=T_{\mathrm{Ours}}-T_{\mathrm{Ref}}
=\left|\widehat{\mathbb E}_{\mathrm{Ours}}[Q_T]\right|
-\left|\widehat{\mathbb E}_{\mathrm{Ref}}[Q_T]\right|.
\label{eq:bootstrapstat}
\end{equation}
A positive value indicates stronger terminal directional bias for Ours, while a negative value indicates greater inventory neutrality. The null hypothesis is
\[
H_0:D\le 0,
\]
i.e., Ours does not have larger terminal directional inventory bias. Under trajectory independence, samples are resampled with replacement within each method. The pooled test uses $B=20{,}000$ resamples and significance level $\alpha=0.05$. A one-sided bootstrap $p$-value of at least $0.05$ is interpreted as no evidence of stronger terminal directional bias for Ours. Under our robustness definition, a method is regarded as more robust when it improves $\CVaR_{10\%}$ without the bootstrap test indicating stronger directional inventory bias.

We also use heatmaps to study how the Hawkes self-excitation parameter $\gamma_H$ and price-impact coefficient $\xi$ jointly affect left-tail robustness.

\begingroup
\scriptsize
\setlength{\tabcolsep}{1.5pt}
\renewcommand{\arraystretch}{1.08}
\begin{longtable}{@{}
>{\raggedright\arraybackslash}p{2.80cm}
>{\raggedright\arraybackslash}p{2.35cm}
>{\raggedright\arraybackslash}p{1.15cm}
>{\centering\arraybackslash}p{2.05cm}
>{\centering\arraybackslash}p{1.25cm}
>{\centering\arraybackslash}p{1.45cm}
>{\centering\arraybackslash}p{1.45cm}
>{\centering\arraybackslash}p{2.05cm}@{}}
\caption{Representative test environments. G1 is the standard environment; G6 and G11 are Hawkes-dominated; G13 and G14 combine high drift and execution pressure; G18 and G19 are extreme/compound regimes. The regime-parameter vector is ordered as $(b,A,k,\kappa_H,\gamma_H,\xi)$. Complete results are reported in the appendix.}
\label{tab:representative}\\
\toprule
Environment & \shortstack[l]{Regime\\parameters} & Method & \shortstack{Terminal\\wealth} & Sharpe & \shortstack{CVaR\\10\%} & \shortstack{CVaR\\30\%} & \shortstack{Terminal\\inventory} \\
\midrule
\endfirsthead
\multicolumn{8}{l}{\tablename\ \thetable\ (continued)}\\[2pt]
\toprule
Environment & \shortstack[l]{Regime\\parameters} & Method & \shortstack{Terminal\\wealth} & Sharpe & \shortstack{CVaR\\10\%} & \shortstack{CVaR\\30\%} & \shortstack{Terminal\\inventory} \\
\midrule
\endhead
\midrule
\multicolumn{8}{r}{Continued on next page}\\
\endfoot
\bottomrule
\endlastfoot

\multirow{4}{2.80cm}{\raggedright \nolinkurl{G1_std}} & \multirow{4}{2.35cm}{\raggedright (0, 140, 1.5, 60, 0, 0)} &
Fixed & 14.9841 $\pm$ 6.9313 & 2.161794 & 3.888963 & 7.313555 & -1.2930 $\pm$ 0.7847 \\
& & Random & 20.9421 $\pm$ 7.1667 & 2.922123 & 9.369169 & 12.935652 & -1.4750 $\pm$ 0.7402 \\
& & ARLMM & 19.7519 $\pm$ 6.9503 & 2.841865 & 8.172405 & 11.832854 & -0.5350 $\pm$ 0.7612 \\
& & Ours & \textbf{27.0037 $\pm$ 7.5168} & \textbf{3.592460} & \textbf{14.633210} & \textbf{18.339891} & -0.1490 $\pm$ 1.0526 \\
\addlinespace[3pt]

\multirow{4}{2.80cm}{\raggedright \nolinkurl{G6_hawkes_only}} & \multirow{4}{2.35cm}{\raggedright (0, 140, 1.5, 60, 20, 0)} &
Fixed & 21.0198 $\pm$ 8.2750 & 2.540161 & 7.424113 & 11.704850 & -1.3000 $\pm$ 0.8029 \\
& & Random & \emph{29.9129 $\pm$ 8.7390} & \emph{3.422921} & \emph{15.247651} & \emph{20.187225} & -1.4360 $\pm$ 0.7553 \\
& & ARLMM & 28.3563 $\pm$ 8.0409 & 3.526515 & 15.146983 & 19.310301 & -0.5350 $\pm$ 0.7506 \\
& & Ours & 39.1673 $\pm$ 8.9248 & 4.388613 & 24.492749 & 29.137957 & -0.1790 $\pm$ 1.0564 \\
\addlinespace[3pt]

\multirow{4}{2.80cm}{\raggedright \nolinkurl{G11_extreme_hawkes}} & \multirow{4}{2.35cm}{\raggedright (0, 140, 1.5, 45, 40, 0)} &
Fixed & 20.8308 $\pm$ 8.0542 & 2.586322 & 7.878976 & 11.922838 & -1.3530 $\pm$ 0.7868 \\
& & Random & 29.7507 $\pm$ 8.6316 & 3.446706 & 15.392203 & 19.971002 & -1.4620 $\pm$ 0.7288 \\
& & ARLMM & 28.5050 $\pm$ 8.0734 & 3.530709 & 14.882185 & 19.137334 & -0.5190 $\pm$ 0.7240 \\
& & Ours & \textbf{39.0256 $\pm$ 8.4942} & \textbf{4.594403} & \textbf{24.291729} & \textbf{29.212230} & \textbf{-0.1470 $\pm$ 1.0614} \\
\addlinespace[3pt]

\multirow{4}{2.80cm}{\raggedright \nolinkurl{G13_extreme_combo_up}} & \multirow{4}{2.35cm}{\raggedright (5, 105, 1.875, 45, 40, 1.0)} &
Fixed & 5.1131 $\pm$ 5.6062 & 0.912035 & -4.944478 & -1.406936 & -1.3640 $\pm$ 0.8137 \\
& & Random & 9.1712 $\pm$ 5.9797 & 1.533708 & -1.415885 & 2.336459 & -1.4730 $\pm$ 0.7055 \\
& & ARLMM & 13.1582 $\pm$ 5.1424 & 2.558757 & 4.465327 & 7.335191 & -0.5480 $\pm$ 0.7670 \\
& & Ours & \textbf{20.0622 $\pm$ 5.4765} & \textbf{3.663299} & \textbf{10.718706} & \textbf{13.809110} & \textbf{-0.1660 $\pm$ 1.0806} \\
\addlinespace[3pt]

\multirow{4}{2.80cm}{\raggedright \nolinkurl{G14_extreme_combo_down}} & \multirow{4}{2.35cm}{\raggedright (-5, 175, 1.125, 45, 40, 1.0)} &
Fixed & 28.3216 $\pm$ 9.3183 & 3.039340 & 13.800967 & 18.080015 & -1.3040 $\pm$ 0.7799 \\
& & Random & 37.6318 $\pm$ 9.7456 & 3.861408 & 21.875180 & 26.418199 & -1.4600 $\pm$ 0.7204 \\
& & ARLMM & 33.2700 $\pm$ 8.9068 & 3.735368 & 18.624493 & 23.485897 & -0.5580 $\pm$ 0.7370 \\
& & Ours & \textbf{42.5229 $\pm$ 9.7747} & \textbf{4.350317} & \textbf{26.418352} & \textbf{31.491854} & \textbf{-0.1570 $\pm$ 1.0543} \\
\addlinespace[3pt]

\multirow{4}{2.80cm}{\raggedright \nolinkurl{G18_ultra_dense_up_hawkes_only}} & \multirow{4}{2.35cm}{\raggedright (5, 175, 1.125, 40, 40, 0.0)} &
Fixed & 22.7404 $\pm$ 10.8924 & 2.087740 & 5.022351 & 10.469081 & -1.3380 $\pm$ 0.8237 \\
& & Random & 33.7389 $\pm$ 11.6118 & 2.905577 & 13.687681 & 20.467093 & -1.4730 $\pm$ 0.7126 \\
& & ARLMM & 36.0455 $\pm$ 10.9279 & 3.298473 & 18.194456 & 23.608191 & -0.5490 $\pm$ 0.7551 \\
& & Ours & \textbf{51.8173 $\pm$ 12.0118} & \textbf{4.313877} & \textbf{31.800296} & \textbf{38.182402} & \textbf{-0.1380 $\pm$ 1.0441} \\
\addlinespace[3pt]

\multirow{4}{2.80cm}{\raggedright \nolinkurl{G19_ultra_sparse_down_hawkes_only}} & \multirow{4}{2.35cm}{\raggedright (-5, 105, 1.875, 40, 40, 0.0)} &
Fixed & 21.8007 $\pm$ 6.7038 & 3.251980 & 10.691281 & 14.192523 & -1.3260 $\pm$ 0.7965 \\
& & Random & 29.8426 $\pm$ 7.3310 & 4.070738 & 18.130170 & 21.546057 & -1.5080 $\pm$ 0.7228 \\
& & ARLMM & 24.7564 $\pm$ 6.9500 & 3.562071 & 13.496312 & 16.927063 & -0.5460 $\pm$ 0.7697 \\
& & Ours & \textbf{31.6350 $\pm$ 7.5567} & \textbf{4.186356} & \textbf{19.220036} & \textbf{23.042527} & \textbf{-0.1490 $\pm$ 1.0918} \\
\end{longtable}
\endgroup

\subsection{Results}
Extending ARLMM to Hawkes self-excitation and price impact produces a substantially more non-stationary simulation environment. Unlike Random, which samples regimes independently, ARLMM adaptively chooses regimes through an adversarial game; the resulting transition kernel can therefore concentrate more strongly around locally unfavorable regions.

\begin{figure}[H]
\centering
\includegraphics[width=0.78\linewidth]{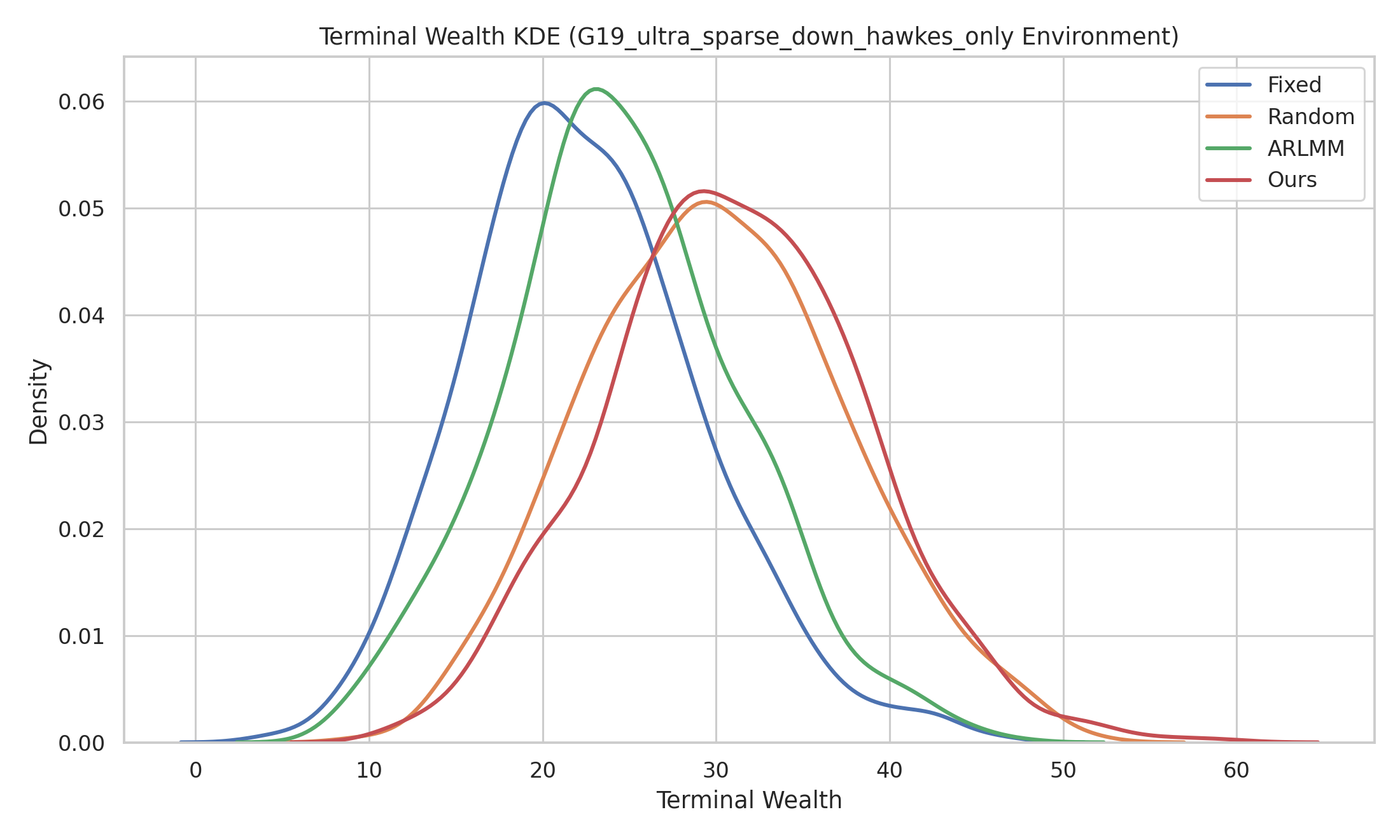}
\caption{Kernel density estimates of terminal wealth in G19.}
\label{fig:kde}
\end{figure}

\paragraph{RQ1: Are left-tail metrics necessary?} Yes, they are informative for the worst-case objective. In the most complex regime, G19, the kernel density estimate of terminal wealth shows clear right-tail expansion for all four methods (Figure~\ref{fig:kde}). A left-tail statistic therefore isolates the part of the distribution most directly targeted by adversarial worst-case training.

\paragraph{RQ2: Is ARLMM consistently more robust than Random in complex environments?} No. According to the primary $\CVaR_{10\%}$ metric, ARLMM outperforms Random in 9 of the 19 environments and is worse in the remaining 10. The underperforming cases cluster into three groups. The first consists of Hawkes-dominated regimes such as G6, G11, and G19. In these environments, risk is driven by clustered order flow and propagates across time steps; a flattened history window has difficulty reliably distinguishing whether a burst is beginning, expanding, or decaying. The second group combines high drift with high execution pressure, such as G10, G14, and G16. Frequent execution accelerates inventory displacement while drift magnifies losses in the unfavorable direction, producing a dynamic chain of ``frequent execution--inventory displacement--further adverse price movement.'' The third group is dominated by liquidity structure, including G1, G4, and G5. Adversarial training tends to concentrate on a small number of unfavorable regimes, whereas Random samples liquidity states more evenly, resulting in smoother training signals and sometimes better generalization.

\paragraph{RQ3: Does the LSTM extension address ARLMM's generalization weakness?} Largely yes. Ours uses the same 40-step history window as ARLMM but applies an LSTM to preserve order and transition patterns. Ours achieves higher $\CVaR_{10\%}$ than ARLMM in all 19 environments. The improvement is consistent across the three problematic classes described above, suggesting that temporal state representation helps identify whether order-flow clustering is intensifying, whether the execution--inventory--price-risk chain is developing, and how liquidity conditions are changing. A static window representation primarily captures what happened recently, while the LSTM can encode how those events are evolving over time.

\paragraph{RQ4: Is Ours more robust across all environments?} Relative to ARLMM, Ours has higher $\CVaR_{10\%}$ in 19/19 environments. Relative to Random, Ours is better in 18/19 environments; the exception is G3\_drift\_down, which contains neither Hawkes clustering nor price impact, so sequential structure provides less additional information. In the pooled inventory test, Ours versus Random gives $D_{\mathrm{obs}}=-1.29968$ with one-sided bootstrap $p=0.504425$, and Ours versus ARLMM gives $D_{\mathrm{obs}}=-0.379579$ with $p=0.500925$. Thus, at the 5\% significance level, the tests do not provide evidence that Ours has stronger terminal directional inventory bias.

\begin{figure}[H]
\centering
\includegraphics[width=0.48\linewidth]{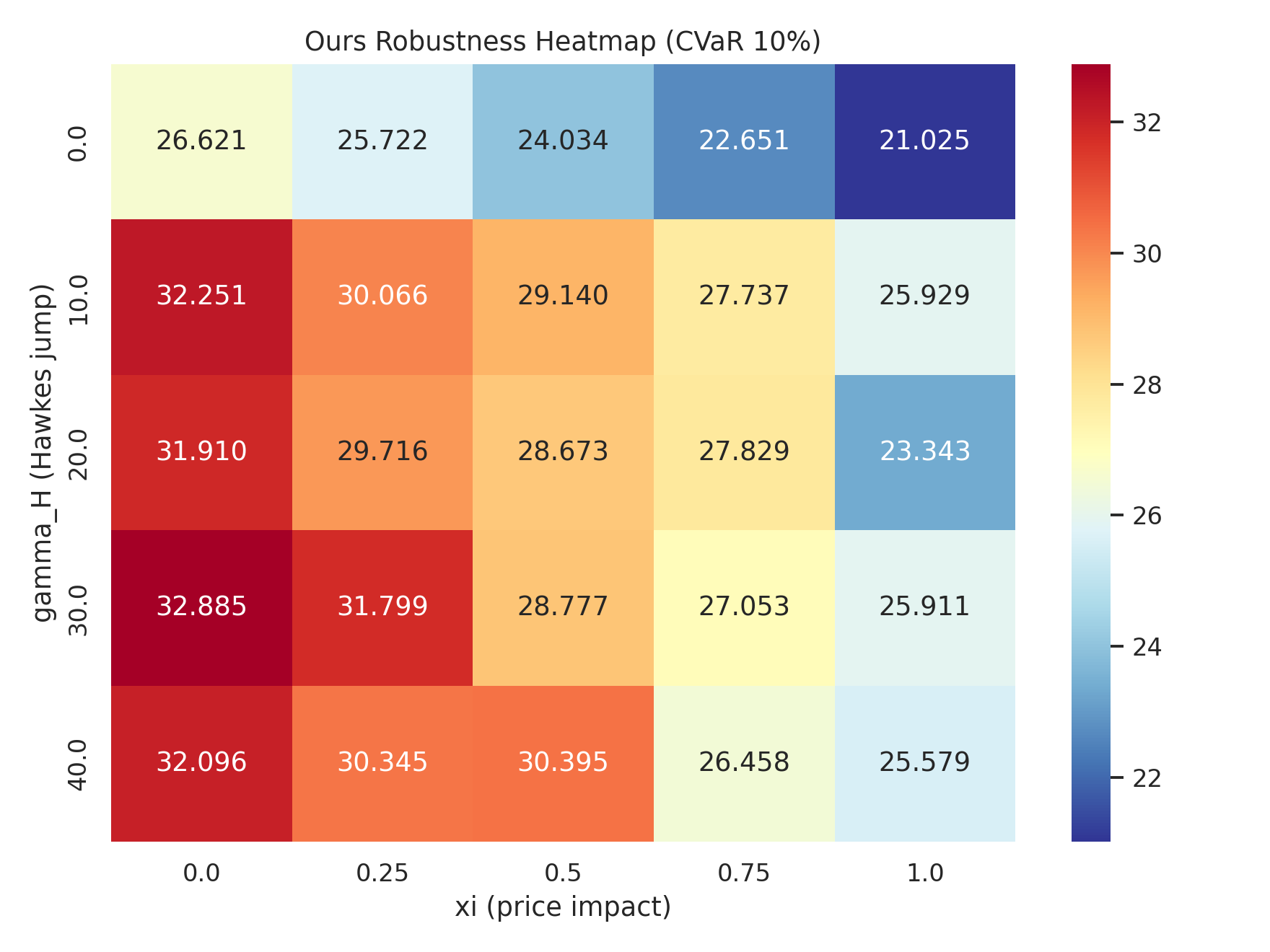}\hfill
\includegraphics[width=0.48\linewidth]{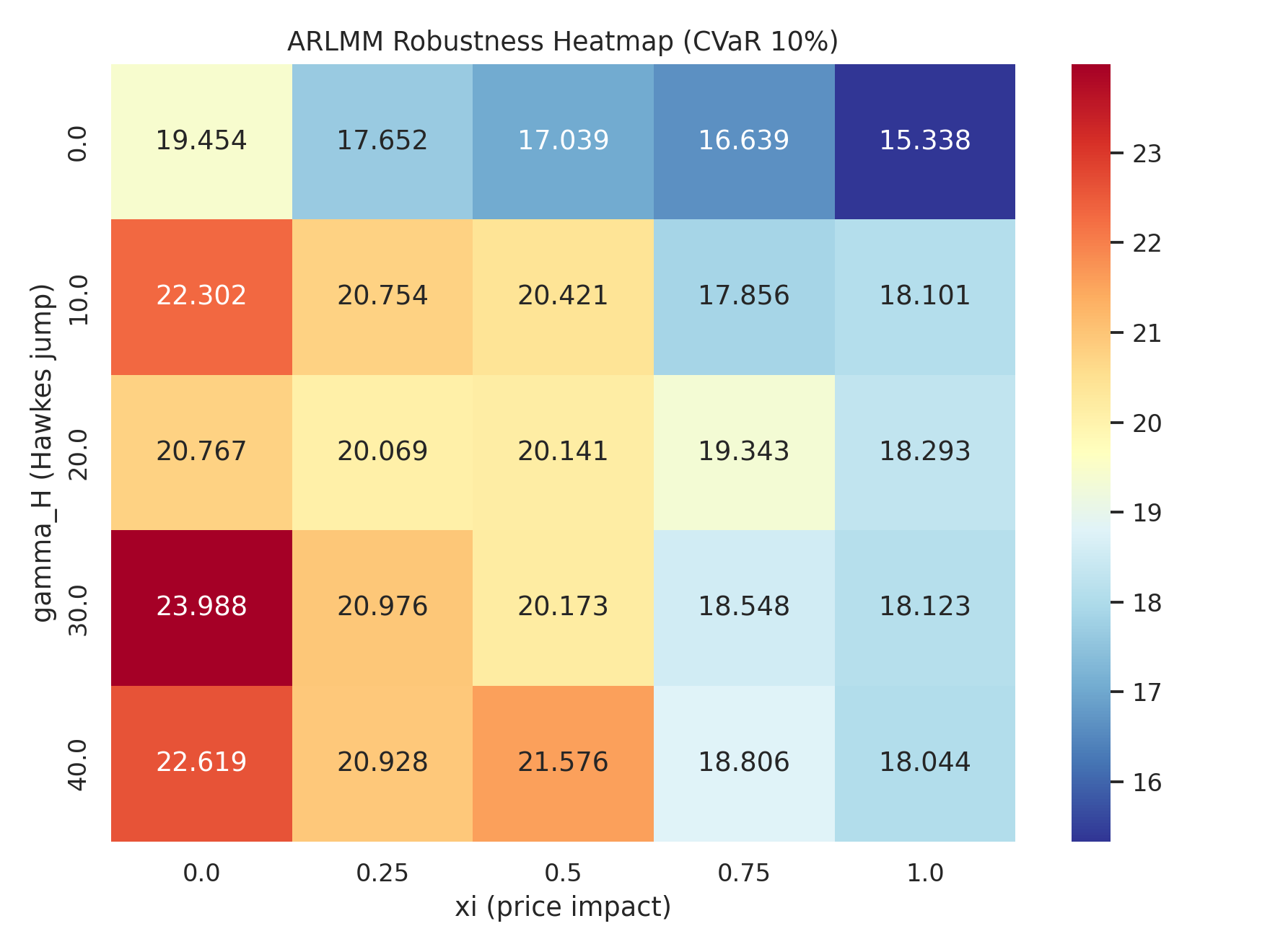}
\caption{$\CVaR_{10\%}$ heatmaps over $(\gamma_H,\xi)$ for Ours and ARLMM.}
\label{fig:heatmap}
\end{figure}

\paragraph{RQ5: In which regimes is the LSTM extension most effective?} Figure~\ref{fig:heatmap} shows that Ours exceeds ARLMM at all 25 grid points in the $(\gamma_H,\xi)$ plane. The largest gains occur around $(\gamma_H,\xi)=(20,0)$ and $(30,0.25)$, while the smallest gain is near $(20,1.0)$. The advantage is largest in medium-to-high Hawkes regimes with low price impact. In such markets, liquidity can build up over several steps before receding, creating cumulative inventory risk whose phase is observable from recent trajectories. The LSTM can compress this temporal information into an internal state. When Hawkes effects coexist with moderate impact, the method remains effective because the policy can jointly infer arrival clustering and the price consequences of recent executions. At very high impact ($\xi=1$), the incremental benefit becomes smaller because a larger fraction of loss comes from the immediate execution itself rather than multi-step risk propagation.

\paragraph{RQ6: Is approximate Nash equilibrium sufficient as a definition of robustness?} No. The true Nash gap in a multi-stage continuous-action game is difficult to compute exactly, so empirical best-response analysis only provides an approximate equilibrium diagnostic. Moreover, low exploitability within the training game does not guarantee strong left-tail performance in unseen markets. Both ARLMM and Ours exhibit approximate-equilibrium behavior, yet ARLMM is still more prone to left-tail deterioration in complex test environments. We therefore treat approximate equilibrium as complementary evidence of game stability rather than as the primary definition of robustness.

\section{Conclusion}
We studied robust market-making policy learning under complex market microstructure. In contrast with the original ARLMM setting, which is based primarily on Poisson arrivals and no direct price impact, our environment includes Hawkes self-exciting order arrivals and trade-induced price impact, thereby representing order-flow clustering, temporal risk propagation, and post-trade price feedback.

The experiments show that, in richer microstructure environments, the original ARLMM is not uniformly superior to randomly sampled regime training. Its weaknesses are concentrated in Hawkes-dominated regimes, environments with high drift and execution pressure, and environments dominated by liquidity structure. Adding an LSTM over the same historical observation window explicitly models sequential structure and improves left-tail performance in nearly all evaluation environments, indicating that temporal state representation is useful for identifying hidden regimes and propagating risk states.

We also analyzed the equilibrium properties of the extended framework in a single-stage reduced-form game and used multi-stage numerical evidence to study training stability. To match the worst-case nature of adversarial training, we used $\CVaR_{10\%}$ as the primary robustness metric and combined it with a one-sided bootstrap test of terminal inventory bias. The test results provide no evidence that the observed left-tail improvement is driven by stronger terminal directional inventory exposure.

Future work can proceed in three directions: extending the single-market-maker model to oligopolistic competition among multiple market makers; applying the framework to data-driven limit-order-book environments and multi-asset market making; and developing stronger convergence guarantees for multi-stage adversarial games with complex microstructure. Overall, the results indicate that combining adversarial training with temporal state modeling can improve the robustness of reinforcement-learning market-making policies under self-exciting order flow, price impact, and regime uncertainty.

\appendix
\section{Derivation of the Single-Stage Payoff}
From Equations~\eqref{eq:cash} and~\eqref{eq:wealth},
\[
\Delta\Pi=\delta^-\Delta N^-+\delta^+\Delta N^+ + H\Delta Z+\Delta H\Delta Z.
\]
Using Equations~\eqref{eq:midprice} and~\eqref{eq:inventory},
\[
\Delta Z=b\Delta t+\sigma W+\xi(\Delta N^- -\Delta N^+),\qquad
\Delta H=\Delta N^+-\Delta N^-.
\]
Hence
\[
\Delta H\Delta Z=(\Delta N^+-\Delta N^-)
\left[b\Delta t+\sigma W+\xi(\Delta N^- -\Delta N^+)\right].
\]
Taking conditional expectations, using $\mathbb E[W]=0$ and, for a small time step,
\[
\mathbb E[\Delta N^{\pm}]=\Lambda^{\pm}\Delta t,\qquad
\mathbb E[(\Delta N^{\pm})^2]=\Lambda^{\pm}\Delta t+o(\Delta t),\qquad
\mathbb E[\Delta N^+\Delta N^-]=o(\Delta t),
\]
gives
\[
\mathbb E[\Delta H\Delta Z]
=b(\Lambda^+-\Lambda^-)\Delta t-\xi(\Lambda^++\Lambda^-)\Delta t+o(\Delta t).
\]
Moreover,
\[
\mathbb E[\delta^-\Delta N^-+\delta^+\Delta N^+]
=(\Lambda^-\delta^-+\Lambda^+\delta^+)\Delta t,
\qquad
\mathbb E[H\Delta Z]=bH\Delta t+o(\Delta t).
\]
Therefore,
\[
\mathbb E[\Delta\Pi]
=\left[\Lambda^+(\delta^+ + b-\xi)
+\Lambda^-(\delta^- - b-\xi)+bH\right]\Delta t+o(\Delta t),
\]
which yields Equation~\eqref{eq:singlepayoff} after removing the common factor $\Delta t$.

\section{Proof of Lemma 1}
For the positive side,
\[
\frac{\partial^2 f}{\partial(\delta^+)^2}
=\lambda^+ e^{-k^+\delta^+}
\left[(k^+)^2(\delta^+ + b-\xi)-2k^+\right].
\]
For the negative side,
\[
\frac{\partial^2 f}{\partial(\delta^-)^2}
=\lambda^- e^{-k^-\delta^-}
\left[(k^-)^2(\delta^- - b-\xi)-2k^-\right].
\]
The mixed derivative is zero, so the Hessian is diagonal. The stated intervals make both diagonal entries non-positive, proving concavity.

\section{Proof of Theorem 1}
\label{app:theorem1}
Fix $(\lambda^{\pm},k^{\pm},\xi)$. By Lemma~1, $f$ is concave in the market maker's quote-depth strategy on the stated compact convex set. For fixed quote depths, Equation~\eqref{eq:singlepayoff} is affine in $b$, and is therefore both convex and concave. Since both strategy sets are nonempty, compact, and convex, Sion's minimax theorem applies, yielding a pure-strategy saddle point and hence a pure-strategy Nash equilibrium.

At an interior equilibrium, the first-order conditions give Equation~\eqref{eq:bestresponse}. Substituting the best response into the payoff defines
\[
\phi(b)=f(\delta^+(b),\delta^-(b);b,\lambda^{\pm},k^{\pm},\xi).
\]
Because $\phi$ is continuous on the compact interval $[\underline b,\overline b]$, a minimizer exists. Any
\[
b^*\in\arg\min_{b\in[\underline b,\overline b]}\phi(b)
\]
completes the equilibrium characterization.

\section{Proof of Theorem 2}
\label{app:theorem2}
Let $D$ and $\Theta$ denote the compact strategy sets of the market maker and the adversary. Because $f$ is continuous on $D\times\Theta$, the sets of probability measures over the two compact strategy spaces are nonempty, compact, and convex. Extending $f$ linearly to mixed strategies gives a continuous bilinear expected-payoff function. By Glicksberg's theorem, the continuous game admits a mixed-strategy Nash equilibrium; in the zero-sum case, this is equivalent to a mixed-strategy saddle point.

\section{Complete Test Results}
\begingroup
\scriptsize
\setlength{\tabcolsep}{1.5pt}
\renewcommand{\arraystretch}{1.08}
\begin{longtable}{@{}
>{\raggedright\arraybackslash}p{2.80cm}
>{\raggedright\arraybackslash}p{2.35cm}
>{\raggedright\arraybackslash}p{1.15cm}
>{\centering\arraybackslash}p{2.05cm}
>{\centering\arraybackslash}p{1.25cm}
>{\centering\arraybackslash}p{1.45cm}
>{\centering\arraybackslash}p{1.45cm}
>{\centering\arraybackslash}p{2.05cm}@{}}
\caption{Complete test results across all 19 evaluation environments. The regime-parameter vector is ordered as $(b,A,k,\kappa_H,\gamma_H,\xi)$.}
\label{tab:complete-results}\\
\toprule
Environment & \shortstack[l]{Regime\\parameters} & Method & \shortstack{Terminal\\wealth} & Sharpe & \shortstack{CVaR\\10\%} & \shortstack{CVaR\\30\%} & \shortstack{Terminal\\inventory} \\
\midrule
\endfirsthead
\multicolumn{8}{l}{\tablename\ \thetable\ (continued)}\\[2pt]
\toprule
Environment & \shortstack[l]{Regime\\parameters} & Method & \shortstack{Terminal\\wealth} & Sharpe & \shortstack{CVaR\\10\%} & \shortstack{CVaR\\30\%} & \shortstack{Terminal\\inventory} \\
\midrule
\endhead
\midrule
\multicolumn{8}{r}{Continued on next page}\\
\endfoot
\bottomrule
\endlastfoot

\multirow{4}{2.80cm}{\raggedright \nolinkurl{G1_std}} & \multirow{4}{2.35cm}{\raggedright (0, 140, 1.5, 60, 0, 0)} & Fixed & 14.9841 $\pm$ 6.9313 & 2.161794 & 3.888963 & 7.313555 & -1.2930 $\pm$ 0.7847 \\*
& & Random & 20.9421 $\pm$ 7.1667 & 2.922123 & 9.369169 & 12.935652 & -1.4750 $\pm$ 0.7402 \\*
& & ARLMM & 19.7519 $\pm$ 6.9503 & 2.841865 & 8.172405 & 11.832854 & -0.5350 $\pm$ 0.7612 \\*
& & Ours & 27.0037 $\pm$ 7.5168 & 3.592460 & 14.633210 & 18.339891 & -0.1490 $\pm$ 1.0526 \\
\addlinespace[3pt]

\multirow{4}{2.80cm}{\raggedright \nolinkurl{G2_drift_up}} & \multirow{4}{2.35cm}{\raggedright (5, 140, 1.5, 60, 0, 0)} & Fixed & 9.8071 $\pm$ 6.8781 & 1.425851 & -1.625236 & 1.927832 & -1.2750 $\pm$ 0.7576 \\*
& & Random & 15.2982 $\pm$ 7.7047 & 1.985568 & 2.642664 & 6.452767 & -1.4040 $\pm$ 0.7108 \\*
& & ARLMM & 17.5741 $\pm$ 7.1273 & 2.465752 & 5.673189 & 9.582654 & -0.5630 $\pm$ 0.7407 \\*
& & Ours & 26.9541 $\pm$ 7.0864 & 3.803657 & 15.202469 & 18.782662 & -0.1970 $\pm$ 1.0841 \\
\addlinespace[3pt]

\multirow{4}{2.80cm}{\raggedright \nolinkurl{G3_drift_down}} & \multirow{4}{2.35cm}{\raggedright (-5, 140, 1.5, 60, 0, 0)} & Fixed & 20.0805 $\pm$ 6.9358 & 2.895208 & 8.966669 & 12.130101 & -1.3250 $\pm$ 0.7681 \\*
& & Random & 27.2630 $\pm$ 7.1865 & 3.793657 & 15.246671 & 19.060111 & -1.4420 $\pm$ 0.7545 \\*
& & ARLMM & 22.1679 $\pm$ 6.9766 & 3.177482 & 10.720030 & 14.282200 & -0.5970 $\pm$ 0.7754 \\*
& & Ours & 27.8852 $\pm$ 8.0087 & 3.481877 & 14.185857 & 18.773974 & -0.1500 $\pm$ 1.0605 \\
\addlinespace[3pt]

\multirow{4}{2.80cm}{\raggedright \nolinkurl{G4_sparse_flow}} & \multirow{4}{2.35cm}{\raggedright (0, 105, 1.875, 60, 0, 0)} & Fixed & 9.2223 $\pm$ 5.3318 & 1.729677 & 0.731933 & 3.274471 & -1.2630 $\pm$ 0.7928 \\*
& & Random & 12.8024 $\pm$ 5.4242 & 2.360233 & 3.987363 & 6.652077 & -1.4050 $\pm$ 0.7180 \\*
& & ARLMM & 11.8479 $\pm$ 4.8907 & 2.422547 & 3.935737 & 6.310026 & -0.4710 $\pm$ 0.7468 \\*
& & Ours & 16.2780 $\pm$ 5.5888 & 2.912632 & 7.087150 & 9.911203 & -0.1480 $\pm$ 1.0655 \\
\addlinespace[3pt]

\multirow{4}{2.80cm}{\raggedright \nolinkurl{G5_dense_flow}} & \multirow{4}{2.35cm}{\raggedright (0, 175, 1.125, 60, 0, 0)} & Fixed & 24.4508 $\pm$ 9.8275 & 2.487981 & 7.949146 & 13.346800 & -1.3130 $\pm$ 0.7769 \\*
& & Random & 35.2973 $\pm$ 10.5594 & 3.342749 & 17.957651 & 23.316411 & -1.5010 $\pm$ 0.6989 \\*
& & ARLMM & 33.6512 $\pm$ 10.1959 & 3.300456 & 17.311388 & 22.396070 & -0.5570 $\pm$ 0.7452 \\*
& & Ours & 45.8813 $\pm$ 10.9520 & 4.189295 & 27.287987 & 33.430723 & -0.1720 $\pm$ 1.0861 \\
\addlinespace[3pt]

\multirow{4}{2.80cm}{\raggedright \nolinkurl{G6_hawkes_only}} & \multirow{4}{2.35cm}{\raggedright (0, 140, 1.5, 60, 20, 0)} & Fixed & 21.0198 $\pm$ 8.2750 & 2.540161 & 7.424113 & 11.704850 & -1.3000 $\pm$ 0.8029 \\*
& & Random & 29.9129 $\pm$ 8.7390 & 3.422921 & 15.247651 & 20.187225 & -1.4360 $\pm$ 0.7553 \\*
& & ARLMM & 28.3563 $\pm$ 8.0409 & 3.526515 & 15.146983 & 19.310301 & -0.5350 $\pm$ 0.7506 \\*
& & Ours & 39.1673 $\pm$ 8.9248 & 4.388613 & 24.492749 & 29.137957 & -0.1790 $\pm$ 1.0564 \\
\addlinespace[3pt]

\multirow{4}{2.80cm}{\raggedright \nolinkurl{G7_impact_only}} & \multirow{4}{2.35cm}{\raggedright (0, 140, 1.5, 60, 0, 0.6)} & Fixed & 12.3512 $\pm$ 6.0300 & 2.048286 & 2.625264 & 5.564356 & -1.3290 $\pm$ 0.8083 \\*
& & Random & 16.7798 $\pm$ 6.5637 & 2.556461 & 6.189049 & 9.347660 & -1.4480 $\pm$ 0.6912 \\*
& & ARLMM & 16.9432 $\pm$ 6.1102 & 2.772912 & 6.821393 & 10.019288 & -0.5750 $\pm$ 0.7463 \\*
& & Ours & 22.5952 $\pm$ 6.4368 & 3.510317 & 12.322769 & 15.407537 & -0.1960 $\pm$ 1.0755 \\
\addlinespace[3pt]

\multirow{4}{2.80cm}{\raggedright \nolinkurl{G8_complex}} & \multirow{4}{2.35cm}{\raggedright (2.5, 120, 1.75, 50, 30, 0.8)} & Fixed & 10.6606 $\pm$ 6.1853 & 1.723535 & 0.472063 & 3.633379 & -1.3470 $\pm$ 0.8020 \\*
& & Random & 15.4678 $\pm$ 6.2851 & 2.461025 & 4.627163 & 8.328559 & -1.5150 $\pm$ 0.7539 \\*
& & ARLMM & 17.1137 $\pm$ 5.8470 & 2.926891 & 7.983051 & 10.830788 & -0.5460 $\pm$ 0.7619 \\*
& & Ours & 25.0330 $\pm$ 6.0675 & 4.125736 & 14.720953 & 18.126400 & -0.1300 $\pm$ 1.0799 \\
\addlinespace[3pt]

\multirow{4}{2.80cm}{\raggedright \nolinkurl{G9_extreme_sparse_drift_up}} & \multirow{4}{2.35cm}{\raggedright (5, 105, 1.875, 60, 0, 0)} & Fixed & 4.1723 $\pm$ 5.3492 & 0.779980 & -4.617788 & -1.853162 & -1.2480 $\pm$ 0.7792 \\*
& & Random & 7.0685 $\pm$ 5.5814 & 1.266424 & -1.812889 & 0.884344 & -1.4490 $\pm$ 0.7127 \\*
& & ARLMM & 9.8539 $\pm$ 5.4042 & 1.823360 & 1.012248 & 3.787512 & -0.5210 $\pm$ 0.7747 \\*
& & Ours & 16.3462 $\pm$ 5.6440 & 2.896186 & 7.112391 & 10.020768 & -0.1890 $\pm$ 1.0772 \\
\addlinespace[3pt]

\multirow{4}{2.80cm}{\raggedright \nolinkurl{G10_extreme_dense_drift_down}} & \multirow{4}{2.35cm}{\raggedright (-5, 175, 1.125, 60, 0, 0)} & Fixed & 30.4900 $\pm$ 10.0853 & 3.023201 & 13.615354 & 18.966228 & -1.3130 $\pm$ 0.7574 \\*
& & Random & 41.0553 $\pm$ 10.3707 & 3.958785 & 24.016479 & 29.380691 & -1.4700 $\pm$ 0.7278 \\*
& & ARLMM & 35.3678 $\pm$ 10.3866 & 3.405146 & 18.383902 & 23.958171 & -0.5380 $\pm$ 0.7856 \\*
& & Ours & 46.0199 $\pm$ 11.3091 & 4.069274 & 27.316544 & 33.317858 & -0.1590 $\pm$ 1.0153 \\
\addlinespace[3pt]

\multirow{4}{2.80cm}{\raggedright \nolinkurl{G11_extreme_hawkes}} & \multirow{4}{2.35cm}{\raggedright (0, 140, 1.5, 45, 40, 0)} & Fixed & 20.8308 $\pm$ 8.0542 & 2.586322 & 7.878976 & 11.922838 & -1.3530 $\pm$ 0.7868 \\*
& & Random & 29.7507 $\pm$ 8.6316 & 3.446706 & 15.392203 & 19.971002 & -1.4620 $\pm$ 0.7288 \\*
& & ARLMM & 28.5050 $\pm$ 8.0734 & 3.530709 & 14.882185 & 19.137334 & -0.5190 $\pm$ 0.7240 \\*
& & Ours & 39.0256 $\pm$ 8.4942 & 4.594403 & 24.291729 & 29.212230 & -0.1470 $\pm$ 1.0614 \\
\addlinespace[3pt]

\multirow{4}{2.80cm}{\raggedright \nolinkurl{G12_extreme_impact}} & \multirow{4}{2.35cm}{\raggedright (0, 140, 1.5, 60, 0, 1.0)} & Fixed & 10.1906 $\pm$ 5.5647 & 1.831311 & 1.155301 & 3.926978 & -1.2790 $\pm$ 0.8135 \\*
& & Random & 14.3660 $\pm$ 5.8078 & 2.473573 & 4.879368 & 7.810048 & -1.4500 $\pm$ 0.7043 \\*
& & ARLMM & 14.5531 $\pm$ 5.2695 & 2.761755 & 5.914822 & 8.581371 & -0.5160 $\pm$ 0.7696 \\*
& & Ours & 19.4694 $\pm$ 5.8655 & 3.319296 & 10.026529 & 12.941124 & -0.1020 $\pm$ 1.1004 \\
\addlinespace[3pt]

\multirow{4}{2.80cm}{\raggedright \nolinkurl{G13_extreme_combo_up}} & \multirow{4}{2.35cm}{\raggedright (5, 105, 1.875, 45, 40, 1.0)} & Fixed & 5.1131 $\pm$ 5.6062 & 0.912035 & -4.944478 & -1.406936 & -1.3640 $\pm$ 0.8137 \\*
& & Random & 9.1712 $\pm$ 5.9797 & 1.533708 & -1.415885 & 2.336459 & -1.4730 $\pm$ 0.7055 \\*
& & ARLMM & 13.1582 $\pm$ 5.1424 & 2.558757 & 4.465327 & 7.335191 & -0.5480 $\pm$ 0.7670 \\*
& & Ours & 20.0622 $\pm$ 5.4765 & 3.663299 & 10.718706 & 13.809110 & -0.1660 $\pm$ 1.0806 \\
\addlinespace[3pt]

\multirow{4}{2.80cm}{\raggedright \nolinkurl{G14_extreme_combo_down}} & \multirow{4}{2.35cm}{\raggedright (-5, 175, 1.125, 45, 40, 1.0)} & Fixed & 28.3216 $\pm$ 9.3183 & 3.039340 & 13.800967 & 18.080015 & -1.3040 $\pm$ 0.7799 \\*
& & Random & 37.6318 $\pm$ 9.7456 & 3.861408 & 21.875180 & 26.418199 & -1.4600 $\pm$ 0.7204 \\*
& & ARLMM & 33.2700 $\pm$ 8.9068 & 3.735368 & 18.624493 & 23.485897 & -0.5580 $\pm$ 0.7370 \\*
& & Ours & 42.5229 $\pm$ 9.7747 & 4.350317 & 26.418352 & 31.491854 & -0.1570 $\pm$ 1.0543 \\
\addlinespace[3pt]

\multirow{4}{2.80cm}{\raggedright \nolinkurl{G15_ultra_sparse_up_combo}} & \multirow{4}{2.35cm}{\raggedright (5, 105, 1.875, 40, 40, 1.0)} & Fixed & 4.9811 $\pm$ 5.6090 & 0.888051 & -5.059921 & -1.410661 & -1.3420 $\pm$ 0.8073 \\*
& & Random & 8.8287 $\pm$ 5.5000 & 1.605210 & -0.633932 & 2.510387 & -1.4600 $\pm$ 0.6935 \\*
& & ARLMM & 13.1582 $\pm$ 5.2736 & 2.495132 & 4.640487 & 7.239850 & -0.5570 $\pm$ 0.7755 \\*
& & Ours & 20.1793 $\pm$ 5.5118 & 3.661076 & 11.580225 & 14.082871 & -0.1340 $\pm$ 1.0494 \\
\addlinespace[3pt]

\multirow{4}{2.80cm}{\raggedright \nolinkurl{G16_ultra_dense_down_combo}} & \multirow{4}{2.35cm}{\raggedright (-5, 175, 1.125, 40, 40, 1.0)} & Fixed & 28.1113 $\pm$ 9.2696 & 3.032623 & 12.908292 & 17.753067 & -1.3700 $\pm$ 0.8208 \\*
& & Random & 36.7359 $\pm$ 9.6207 & 3.818437 & 21.562117 & 26.214431 & -1.4480 $\pm$ 0.7320 \\*
& & ARLMM & 33.3068 $\pm$ 9.5280 & 3.495672 & 18.037395 & 22.645817 & -0.5340 $\pm$ 0.7742 \\*
& & Ours & 41.4953 $\pm$ 9.7668 & 4.248624 & 24.779776 & 30.295868 & -0.1010 $\pm$ 1.0927 \\
\addlinespace[3pt]

\multirow{4}{2.80cm}{\raggedright \nolinkurl{G17_ultra_sparse_mid_full_stress}} & \multirow{4}{2.35cm}{\raggedright (0, 105, 1.875, 40, 40, 1.0)} & Fixed & 11.0098 $\pm$ 5.1733 & 2.128202 & 2.516506 & 5.302143 & -1.3120 $\pm$ 0.8242 \\*
& & Random & 15.1550 $\pm$ 5.6037 & 2.704441 & 5.835442 & 8.847402 & -1.4900 $\pm$ 0.7283 \\*
& & ARLMM & 15.4633 $\pm$ 4.9423 & 3.128797 & 7.008117 & 9.779481 & -0.5350 $\pm$ 0.7585 \\*
& & Ours & 20.7177 $\pm$ 5.2625 & 3.936819 & 12.065109 & 14.796100 & -0.1890 $\pm$ 1.0317 \\
\addlinespace[3pt]

\multirow{4}{2.80cm}{\raggedright \nolinkurl{G18_ultra_dense_up_hawkes_only}} & \multirow{4}{2.35cm}{\raggedright (5, 175, 1.125, 40, 40, 0.0)} & Fixed & 22.7404 $\pm$ 10.8924 & 2.087740 & 5.022351 & 10.469081 & -1.3380 $\pm$ 0.8237 \\*
& & Random & 33.7389 $\pm$ 11.6118 & 2.905577 & 13.687681 & 20.467093 & -1.4730 $\pm$ 0.7126 \\*
& & ARLMM & 36.0455 $\pm$ 10.9279 & 3.298473 & 18.194456 & 23.608191 & -0.5490 $\pm$ 0.7551 \\*
& & Ours & 51.8173 $\pm$ 12.0118 & 4.313877 & 31.800296 & 38.182402 & -0.1380 $\pm$ 1.0441 \\
\addlinespace[3pt]

\multirow{4}{2.80cm}{\raggedright \nolinkurl{G19_ultra_sparse_down_hawkes_only}} & \multirow{4}{2.35cm}{\raggedright (-5, 105, 1.875, 40, 40, 0.0)} & Fixed & 21.8007 $\pm$ 6.7038 & 3.251980 & 10.691281 & 14.192523 & -1.3260 $\pm$ 0.7965 \\*
& & Random & 29.8426 $\pm$ 7.3310 & 4.070738 & 18.130170 & 21.546057 & -1.5080 $\pm$ 0.7228 \\*
& & ARLMM & 24.7564 $\pm$ 6.9500 & 3.562071 & 13.496312 & 16.927063 & -0.5460 $\pm$ 0.7697 \\*
& & Ours & 31.6350 $\pm$ 7.5567 & 4.186356 & 19.220036 & 23.042527 & -0.1490 $\pm$ 1.0918 \\
\end{longtable}
\endgroup

\end{document}